# Selective Augmentation: Improving Universal Automatic Phonetic Transcription via G2P Bootstrapping

**Tobias Bystrich[1,2], Julia M. Pritzen[1], Christoph A. Schmidt[1], Claudia Wich-Reif[2]**

[1]Fraunhofer Institute IAIS, Sankt Augustin, Germany
[2]University of Bonn, Bonn, Germany
tobias.bystrich@mailbox.org, {julia.pritzen, christoph.andreas.schmidt}@iais.fraunhofer.de,
claudia.wich-reif@uni-bonn.de

**Abstract**

In the field of universal automatic phonetic transcription (APT), clean and diverse training transcriptions are required. However, such high-quality data is limited. We propose the bootstrapping approach Selective Augmentation to improve the available training transcriptions by selectively transferring distinctions between languages. Based on the model MultIPA, we exemplarily show that we could increase the accuracy of an existing feature (plosive voicing) and add a new feature (plosive aspiration) by augmenting the existing training data using information from a separate helper language (Hindi). We describe intrinsic challenges of the evaluation and develop objective metrics to determine the success: Voicing accuracy was increased by 17.6% by reducing the number of false positives. Additionally, aspiration recognition was introduced: While the baseline transcribed 0% of German /p, t, k/ as aspirated, our approach transcribed them as aspirated in 61.2% of the cases. Introducing aspiration recognition to APT models allowed for the tenuis class to be successfully reduced by 32.2%, which also reduces the conflations between the test language's plosives.

## 1. Introduction

Phonetic transcription is a more precise and language-independent transcription of the pronunciation of speech than orthography. As Taguchi et al. (2023) state, thousands of languages are threatened, and revival efforts depend on phonetic documentation. Phonetic transcription is also crucial for language education, mispronunciation detection, accent reduction and efficient training of linguistic experts. However, accurate manual phonetic transcription is highly costly due to the time requirements and needed expertise. Automatic phonetic transcription (APT) can massively reduce this cost.

This works seeks to improve APT models such as the Wav2Vec-2.0-based (Baevski and Zhou, 2020) "MultIPA" (Taguchi et al., 2023). To solve the main limitation of clean, phonetically exact and language-independent transcriptions, we propose a bootstrapping approach using helper models: *Selective Augmentation*.

We exemplarily improve a MultIPA-based model by selectively enhancing the accuracy of plosive voicing recognition and adding plosive aspiration recognition. These improvements are relevant since phonation may be challenging to train from G2P sources which are modeled within word borders, and since aspiration is essential to languages such as Chinese languages (Lee and Zee, 2003; Zee, 2015) and Hindi (Ohala, 2015), and crucial for English (Ladefoged, 2015) and German (Kohler, 2015).

We investigated the accuracy of our approach with an acoustic phonetic analysis and propose metrics to objectively measure the success. The approach can be repeated to clean and enhance noisy multilingual data by using the multilingual data itself. Table 1 and 2 show our notation and terminology. Our main contributions are summarized as follows:

- We propose a bootstrapping procedure to clean and enhance transcriptions for APT training
- We augment a MultIPA-based model, improving plosive voicing recognition and achieving aspiration recognition
- Models with necessary segment lists are publicly available via Huggingface (see 6. Conclusion)

| Ortho-graphy | Phonetic Character | Phoneme | Phone |
|---|---|---|---|
| ⟨g⟩ | ⟨[g]⟩ | /g/ | [g] |

Table 1: Notation in this work

| | [-spread glottis] | [+spread glottis] |
|---|---|---|
| **[-voiced]** | Tenuis | Aspirated |
| **[+voiced]** | Voiced | Breathy (voiced) |

Table 2: Phonation types in this work. We use *tenuis* for unaspirated and unvoiced sounds, not to be confused with *lenis* despite some overlap.

## 2. Related Work

The previously named use cases for universal APT require high language-independent accuracy, phonetic as opposed to a language-specific phonemic transcription[1], and limiting language-specific phonetic transcription conventions, e.g., ⟨[mɔːlʌk]⟩ for Southern England English [mɔːlɐk]. The transcriptions and models use the International Phonetic Alphabet (IPA) (International Phonetic Association, 2015), a phonetic alphabet designed to be universal.

Due to the costs of precise manual transcriptions, MultIPA, like earlier models (Taguchi et al., 2023; Xu, Baevski and Auli, 2022), takes its transcriptions from G2P predictions. Taguchi et al. achieved better performance than previous models by carefully selecting language-G2P pairs that allow for relatively accurate phonetic transcriptions. This way, they used much fewer languages than other models but achieved higher performance. They found that clean data is highly important and that limiting each language to only 1 000 segments worked best. The careful selection of G2Ps by Taguchi et al. is perhaps the most significant difference to the previous models which used G2Ps of uncertain quality. Optimal G2Ps are limited. Our approach may be used to overcome the problems of flawed G2Ps or incorporate only the most high-quality information. This is in-line with Taguchi et al.'s finding that clean phonetic transcriptions are highly important for training.

MultIPA – being based on `wav2vec2-large-xlsr-53` (Conneau et al., 2021) – uses CTC which learns approximate temporal alignments for the phones (Graves et al., 2006). Our work uses this built-in alignment to obtain aligned predictions of high-quality predictions.

For pronunciation and spelling, the necessary linguistic information is readily available for many languages. Open information can be found on wiki projects like Wikipedia. In this work, we use primary sources such as the Illustrations of the IPA (cf. Kohler, 2015; cf. Ohala, 2015).

To improve plosive phonation recognition as described above, we will measure the plosives' voice onset time (VOT) (Lisker and Abramson, 1964). VOT is a way to measure voicing and aspiration of plosives with a long tradition (Abramson and Whalen, 2017). Voiced plosives have negative VOT while aspirated plosives have high positive VOT.

## 3. Method

While improving upon MultIPA could be continued with the costly approach of using considerably more languages with few segments, this is likely to cause an issue where distinctions made in only a few languages would be diluted and where specific inaccuracies would remain in specific contexts or even be reinforced.

For example, glottal stops occur in many languages in several situations. However, only Maltese among the training languages represents glottal stops with a character, which leads to a lack of glottal stop transcriptions for most languages, diluting the useful information from Maltese. Conversely, the common omission of glottal stops will be reinforced by taking even more training languages unless this rare information is increased too.

To prevent this information dilution, we propose augmenting the training data of the other six languages with the most reliable information from one language. Apart from applying information from languages within the training data, it can add new phonetic distinctions to the transcriptions which were originally not transcribed in any of the languages. Selective Augmentation may thus create the exact high quality transcription data that conventional APT is lacking.

In this paper, we use Selective Augmentation to improve plosive voicing recognition and add plosive aspiration recognition to a MultIPA model. As a helper language, we chose Hindi since it has all four combinations of the voicing and aspiration distinction (visible in Table 2) and also 5 places of articulation (PoAs) with this distinction (Ohala, 2015). Since Hindi has breathy plosives [ibid.], we additionally augmented breathiness using ⟨[ʱ]⟩ because it is a possible improvement with no extra cost. However, we make no claims about breathy voice, as it is rare (Abramson and Whalen, 2017) and since measuring breathy voice requires a more complex measure (Davis, 1994).

### 3.1 Models

Selective Augmentation depends on a way to align the phonetic information from the helper model (HM) to the original baseline transcriptions. This paper generates alignments by introducing a new model, the reference model (RM). The HM and RM predictions are aligned via CTC timestamps. For comparability, we use RM predictions for training both the baseline model (BM) and target model (TM).

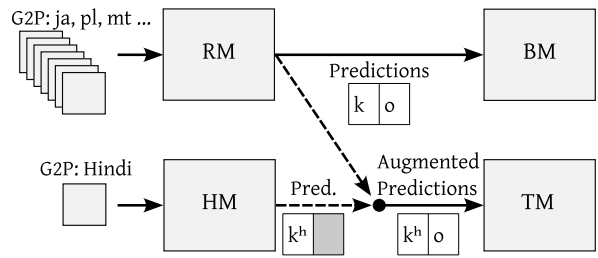

Figure 1: Augmentation procedure and models

---
[1] For the distinction of the terms *phone* vs. *phoneme* compare Moore & Skidmore (2019).

Figure 1 schematically shows the procedure and models used. All models are based on the same training procedure and Wav2Vec 2.0 architecture, only differing in training data. While this work's RM is the direct reproduction of Taguchi et alii's (2023) MultIPA, Selective Augmentation uses the RM for its aligned transcriptions of its own training recordings. The resulting models BM and TM are trained on the model predictions instead of G2P predictions to make them comparable. By the similar usage of RM and HM, we minimize the complexity of the alignment. Using the RM instead of forced alignment is an indirect training step because BM does not see the original G2P output. If Selective Augmentation is effective, we expect that multiple HMs could offset potential inaccuracies caused by our indirect training approach. The only difference between the models is the training data.

### 3.2 Mapping

Table 3 shows which RM and HM predictions can match. The matches are largely based on phonetic feature distance using the IPA feature set, so that ⟨[t̪]⟩ can match with ⟨[t]⟩, but not with ⟨[k]⟩. We also reviewed the HM outputs, which influenced our mapping of HM ⟨[c, ɟ]⟩, corresponding to Hindi /tʃ dʒ/ (cf. Ohala, 2015). The matches are made only when they occur at similar timestamps.

| RM prediction | HM predictions |
|---|---|
| ⟨[p]⟩, ⟨[b]⟩ | ⟨[p, b]⟩ |
| ⟨[t]⟩, ⟨[d]⟩ | ⟨[t, t̪, c, d, d̪, ɟ]⟩ |
| ⟨[t̪]⟩, ⟨[d̪]⟩ | ⟨[t̪, t, d̪, d]⟩ |
| ⟨[k]⟩, ⟨[g]⟩ | ⟨[k, c, g, ɟ]⟩ |
| ⟨[c]⟩, ⟨[ɟ]⟩ | ⟨[c, k, ɟ, g]⟩ |

Table 3: Potentially matching predictions

Since CTC aligns characters approximately and the transcriptions for MultIPA and Hindi are notably different, we needed to determine how precisely the models are aligned temporally. By looking at the respective phone alignments ($i,j$) between RM and HM, we found the alignments to usually be either exact ($i=j$) or one timestamp apart ($j \in [i-1, i+1]$). After investigating the mapping outputs and changes in total mapping count, we further limited the mappings to $[i, i+1]$. The augmentation occurs as follows: If matched, the phonation type of $RM[i]$ is overwritten completely by $HM[j]$, but the PoA is taken purely from the $RM[i]$.

### 3.3 Data

This work builds upon MultIPA, so we used the same training languages for the RM: Finnish, Hungarian, Japanese, Maltese, Modern Greek, Polish and Tamil. We randomly selected 1 000 segments for each language from the *Mozilla Common Voice Corpus 11* (CV 11) (Ardila et al., 2020) for the training split. For the HM, we used 7 000 Hindi segments from CV 11 to match the MultIPA data. While we created a validation split of 5% for the HM data, the RM data uses a validation split of 20%, in line with MultIPA.

To measure improvements beyond the training data accuracy, we conduct an acoustic phonetic analysis using VOT. For this, we needed to use a test dataset with plosives which are voiced, voiceless and aspirated. German is a language for which G2Ps cannot reliably predict the phonetic voicing and aspiration. This made German a suitable testing language. Table 4 visualizes the relationship between the phonemes and phones for German: /p t k/ and /b d g/ both allow for tenuis allophones according to our usage in Table 2 (Kohler, 2015; Becker, 2012: p. 22). While the influential northern pronunciation (cf. ibid.) has been described with voiced /b d g/, Jessen and Ringen (2002) describe that most German dialects show only passive voicing. The trancription ⟨[p t k]⟩ is in language-independent IPA and may, therefore, differ from expected German notation conventions. We also consider only three different realizations for the homorganic phoneme pairs.

| Phonemes | /p t k/ | /b d g/ | |
|---|---|---|---|
| Phones | [pʰ tʰ kʰ] | [p t k] | [b d g] |
| Phonation | Aspirated | Tenuis | Voiced |

Table 4: Relation phoneme to phone for German plosives

We based the German test set on CV to make it diverse and as similar as possible to the MultIPA training data. However, since other research included models also trained on German CV data, such as Wav2Vec2Phoneme (Xu, Baevski and Auli, 2022), we used the official CV delta-segments to control for disjoint training and test data. To ensure that VOT measurements can be as exact as possible, we created a test set with only segment-initial plosives. These will be referred to as *absolute onset realizations*. Most importantly, this ensures an exact distinction between voiced and voiceless closures by avoiding any preceding voicing traces. For this, we used the CV delta-splits 12-14,16-18, while we reserved delta-split 15 for a high-level overview and sample sentences. To pre-filter the data for the absolute onset set, we only considered segments with initial ⟨b d g p t k⟩ in their lower-case transcriptions and evaluated only valid /b d g p t k/. The final dataset contains 40 randomly selected valid and analyzable instances of each phoneme.

### 3.4 Preprocessing

After excluding downvoted segments, the audio data was resampled to 16 kHz and transcribed with G2P models. For Hindi, we used the Python `phonemizer` package (v. 3.2.1), which had some inaccuracies but helped filter out code-switching to English. We manually checked the G2P and found it accurate regarding plosive phonation. For the other data, we used the MultIPA G2P resources.[2]

For every G2P, we replaced all cases of ⟨[ɡ]⟩ with ⟨[g]⟩ to ensure consistency – a consonant which will be affected by our approach. The Tamil dataset had an issue of an unclosed quotation mark, and its large size needed to be addressed to avoid crashes. Finally, a few relatively frequent invalid G2P outputs such as ⟨[éɪ̃]⟩ and remaining orthographic outputs needed to be remapped or the segments excluded.

### 3.5 Metrics

In this language-independent context, we used negative VOT values as an objective measure of voicing. Hence, voicing accuracy (*VoicingAcc*) for /b d ɡ/ can be measured objectively in the described test set.

The voiceless plosives have positive VOT, i.e. voicing lag. However, aspiration is language dependent because the border between tenuis and aspirated can lie at different VOTs (Lisker and Abramson, 1964), creating an area of ambiguity. Hence, we developed metrics that judge the improvement in an objective way.

Since we investigate the plosive realizations on German, a usage-oriented approach may intend to separate the phonemes among /p t k/ realizations for the following reasons: We expect aspiration for /p t k/ in the absolute onset in standard or northern pronunciations (Kohler, 2015; Becker, 2012: p. 24). However, we cannot assume that all German /p t k/ should be transcribed as aspirated because aspiration of these phonemes differs by factors such as environment and spontaneous stress [ibid.]. In addition, regional accent speakers should not be excluded. This means that the aspiration decision border for aspiration should be among /p t k/ and not /b d ɡ/. With these considerations, two further numeric metrics are proposed for this case:

We introduce the aspiration percentage (*Asp%*) for /p t k/, for which higher values are better, given that the independent languages limit overfitting. For Asp% no known optimal value exists, as due to class overlap and German variation, we expect it to be below 100%. Ideally, the classes should not overlap more than the phoneme borders. Failing to transcribe most of /p t k/ as aspirated would be considered a poor result.

The tenuis percentage (*Ten%*) for /p t k b d ɡ/ represents the conflation class ⟨[p t k]⟩. From a usage-oriented standpoint and considering German phonology, it is relevant to keep phoneme conflations low (see Table 4), since not all conflations in transcriptions represent true mergers in pronunciation. Therefore, for this metric, lower values are better.

If a model predicts an incorrect active articulator or manner, we define them as NULL predictions which will not influence the metrics. For these, a lower value is better.

Another consideration is that some cases may legitimately be represented as either aspiration or another voiceless (homorganic) continuant, e.g., [kʰu~khu~kxu] or [t͡ʃ]. In a worst-case analysis, the findings should hold even when all ambiguous alternative representations are treated as ⟨[ʰ]⟩. Therefore, we provide a main result which excludes these ambiguous cases, and an additional result in parentheses that treats all ambiguous cases as valid representations.

### 3.6 Limitations

We investigate a subset of the features and phones with only one language, but this is addressed by having an independent testing language and proposing objective metrics. To achieve measurements of the highest objectivity, only plosives in the absolute onset are measured.

## 4. Experiments

The pre-trained `wav2vec2-large-xlsr-53` was retrieved from Huggingface transformers.[3] Reusable MultIPA functions were taken from the Git repository.[3] We used `Wav2Vec2CTCTokenizer` to load the public MultIPA tokenizer with its large IPA vocabulary. To allow for better error detection, we removed ⟨[ɡ]⟩ and other characters known to be unused from the `vocab.json`. This allowed for adding ⟨[ɦ]⟩ which can theoretically be added by the Hindi HM.

### 4.1 Training

Unless explicitly noted, the training parameters are the same as MultIPA. We trained all models with a learning rate of $3 \cdot 10^{-4}$ and 500 warm-up steps, freezing the feature extractor. The remaining parameters are determined by the `transformers` library version 4.26.0. Taguchi et al. trained the corresponding model for approximately four hours on four GTX 1080Ti GPUs, however, this involved 30 epochs of training (Taguchi et al., 2023). We have not found an indication of an early stopping procedure. For our training, we found that ten epochs for 7 000 segments were sufficient to reach a validation optimum (see Table 5). While multiple models were trained for this work, the lowered epochs reduced the costs.

---
[2] https://github.com/ctaguchi/multipa

[3] https://huggingface.co/facebook/wav2vec2-large-xlsr-53

| RM | HM | BM | TM |
|---|---|---|---|
| 3.54 | 9.94 | 5.71 | 5.71 |

Table 5: Epochs at which early stopping was reached

We trained on an HPC cluster (see Section 7), keeping the batch size at 4, as in the original MultIPA. In contrast to Taguchi et al. (2023), the memory was sufficient so that we did not need to remove long segments.

## 4.2 Results

Table 6 shows sample predictions of the MultIPA-based models. We list RM predictions for comparison to the two indirectly trained models.

|    | *Es gibt viel zu tun* | *Ganz toll* | *Danke dafür* |
|---|---|---|---|
| TM | ɛskitfil͡tsutʰun | kan͡tsstʰvɛ | dankʰɛdafɪ |
| BM | ɛskɛtfil͡tsuttʃuːn | kan͡tsstvɛ | dankɛdafɪ |
| RM | ɛskipfil͡tsut͡ʂun | kɛnstvɛ | daŋkɛdafy |

Table 6: Sample predictions

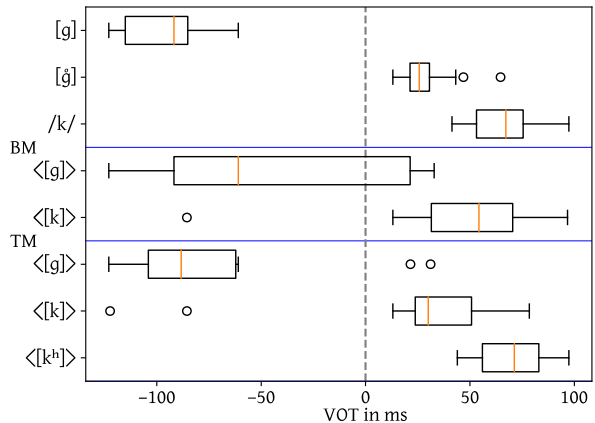

Figure 2: VOT boxplot of velar plosives

Figure 2 exemplarily shows a boxplot of the VOT of the velar plosives. Additional boxplots and tables can be found in the Appendix. In the first section, the three classes are the voiced and unvoiced allophones of /g/ and the phoneme /k/. The model outputs are divided into the investigated phonation types. The two classes in the baseline are successfully split into three classes. The new aspirated class ⟨[kʰ]⟩ closely resembles the distribution of /k/, while the refined voiced class ⟨[g]⟩ has reduced false positives. The tenuis class ⟨[k]⟩ shows one new voiced outlier and is not optimal due to its overlap with ⟨[kʰ]⟩.

Table 7 shows improvements for TM in all cases except for an insignificant increase in NULL predictions, representing a single prediction. The result 0.0% instead of N/A implies that the baseline should support this output given its inclusion in the original `vocab.json` and its scope as a universal APT model, but fails to do so. The values in parentheses show that this holds even in a worst-case analysis with all ambiguous alternative representations (see Section 3.5). Given the small test set for /b d g/, we statistically tested the voicing recognition improvement of 17.6% and found a significance at the 5% level. 30.4% of /b d g/ in the data were found to be voiced. The remaining values are discussed in Section 5.

|    | Voicing Acc | Asp% | Ten% | NULL |
|---|---|---|---|---|
| BM | 71.3 | 0.0 (13.7) | 73.8 (68.4) | 9.2 |
| TM | 83.8 | 61.2 (63.6) | 50.0 (48.4) | 9.6 |

Table 7: Results for all PoAs in percent with ambiguous cases in parentheses. While high Asp% are preferable, for Ten%, lower is better. BM did not recognize aspiration following a strict definition, whereas ambiguous cases may be considered for a value of up to 13.7%.

## 5. Discussion

The results show substantial improvements for the suggested metrics. Despite the successful reduction in false positives for voicing, the tenuis class was simultaneously successfully reduced by 32.2%, which avoids conflations and thereby improves the model further for the test language.

The increase of the Asp% to 61.2% of /p t k/ indicates a great success for the first application of Selective Augmentation. We identified two shortcomings: Firstly, some aspirates were not mapped, as shown by the high upper cap of the tenuis box, causing a significant overlap. The overlap indicates that not all aspirates were successfully mapped and that the mapping procedure may still be improved.

Secondly, /p/ greatly underperforms with an Asp% of only 25.8%. We investigated possible reasons for this and found that, in the training data, ⟨[pʰ]⟩ was mapped only 146 times while ⟨[tʰ]⟩ and ⟨[kʰ]⟩ occurred 1 423 and 1 109 times respectively. This difference and the effect on /p/ suggest that there may be too few segments with significantly aspirated [p] among the training languages. It seems plausible that, for languages with low average VOT, most segments may not contain significant aspiration. Considering the difference observed for fewer aspiration occurrences, increasing the frequency of recordings with significant aspiration could improve the class overlap for all plosives. We could raise the aspiration frequency in a test by pre-selecting 6 345 out of 7 000 training recordings where the existing RM and HM successfully mapped ⟨[ʰ]⟩. However, for optimal aspiration recognition, Hindi may be added to the RM training data after cleaning the data with Selective Augmentation.

It may seem surprising that many pre-voiced plosives (30.4% of /b d g/) were encountered in the German data. This may be due to the careful speech of the speakers, regiolects or sociolects with more frequent onset voicing and occasionally due to non-native speakers.

## 6. Conclusion

The proposed bootstrapping approach Selective Augmentation was used to refine the plosive voicing recognition and add aspiration recognition to a MultIPA based model. For our German test set, we were able to increase voicing accuracy by relative 17.6% and decrease the conflation class by relative 32.2% while increasing the aspiration recognition from 0% to 61.2%. This suggests that Selective Augmentation can be used to both clean and enhance the training transcriptions and thereby address the issue of transcription quality in APT training data. Our results for bilabials suggest supplementing more aspirated training audio could increase aspiration recognition even more.

This work's metrics were developed with great care to evaluate the resulting model against all valid usages of the IPA. For this, all ambiguous cases were also considered. While these metrics may be difficult for non-linguists to develop, they show the high stability of the improvement even compared to phonetic measurements. The strong improvement shown and lack of regressions suggest that one can have high confidence in the model generalizing well to other manners of articulation, particularly since the linguistic evaluation was conducted independently of any test set.

Selective Augmentation can be scaled up for a broad coverage of each language's inaccuracies and thereby create a clean dataset with all IPA distinctions. Future work may apply Selective Augmentation to other ASR approaches like Zipa-Cr (Zhu et al., 2025), apply it to new features or improve the plosive aspiration further by up-sampling aspirated bilabials, which could be done by using helper models to pre-filter the dataset for recognized aspiration.

The phonetic ASR models trained for and mentioned in this work are available via Huggingface, as well as the unique references to the used Common Voice segments used for training.[4]

---

[4]https://huggingface.co/collections/Tobias-B/universal-phonetic-asr-models-selective-augmentation

## 7. Acknowledgments

Simulations were performed with computing resources granted by WestAI under project rwth1594.

Also, we want to acknowledge Chihiro Taguchi since the release of MultIPA and the corresponding GitHub repository resulted in an unexcepted positive impact on our phonetic ASR research.

# Appendix

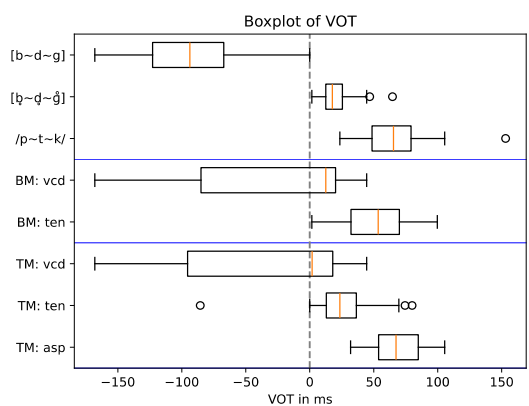

Appendix 1: combined VOT boxplot of all PoAs (corresponds to Table 7)

|    | Voicing Acc | Asp% | Ten% | NULL |
|----|---|---|---|---|
| BM | 78.1 | 0 (5.1) | 78.3 (76.1) | 11.3 |
| TM | 91.9 | 66.7 | 52.6 | 5 |

Appendix 2: Results for velar plosives

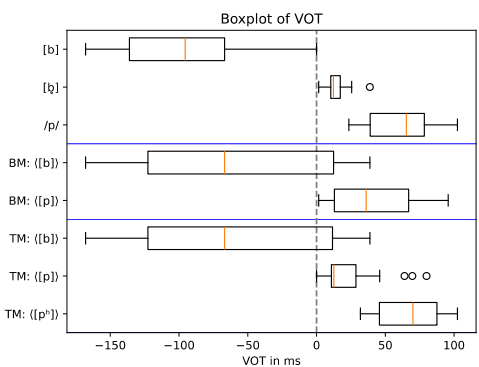

|    | Voicing Acc | Asp% | Ten% | NULL |
|----|---|---|---|---|
| BM | 84.6 | 0 (12.8) | 74.0 (69.2) | **2.5** |
| TM | 94.7 | **25.8 (30.3)** | **66.7 (64.8)** | 7.5 |

Appendix 3: Plot and results for bilabial plosives

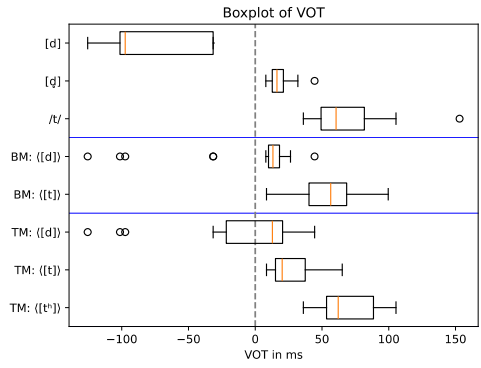

| | Voicing Acc | Asp% | Ten% | NULL |
|---|---|---|---|---|
| **BM** | 46.7 | 0 (23.1) | 68.3 (59.4) | **13.8** |
| **TM** | **60** | **87.9 (89.5)** | **28.6 (26.5)** | 15 |

Appendix 4: Plot and results for alveolar plosives